\documentclass{article}

\usepackage{footnote}
\usepackage{arxiv}
\usepackage{graphicx}
\usepackage{epsfig}
\usepackage{amssymb}
\usepackage{amsmath}
\usepackage{amsfonts}
\usepackage{tabularx}
\usepackage{todonotes}
\usepackage{multirow}
\usepackage{colortbl}
\usepackage{booktabs}
\usepackage{adjustbox}
\usepackage{xspace}
\usepackage{subfigure}
\usepackage{caption}

\usepackage[noadjust]{cite} 

\makeatletter
\let\NAT@parse\undefined
\makeatother
\usepackage{hyperref}

\newcommand{\cl}{\cellcolor[rgb]{ .949,  .949,  .949}}

\newcommand{\clg}[1]{{\color[HTML]{22A884}#1}} 

\newcommand{\survey}[1]{\textit{Survey #1}} 

\newcommand{\smaller}[1]{\fontsize{7pt}{0.1em}\selectfont \ensuremath#1}

\makesavenoteenv{tabular}
\makesavenoteenv{table}
\makeatletter

\title{Tackling Face Verification Edge Cases: In-Depth Analysis and Human-Machine Fusion Approach}

\author{\href{https://orcid.org/0000-0002-0503-4600}{\includegraphics[scale=0.06]{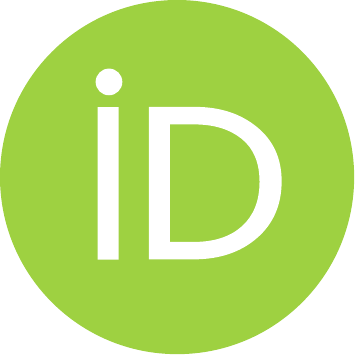}\hspace{1mm}Martin Knoche} \qquad \href{https://orcid.org/0000-0003-1096-1596}{\includegraphics[scale=0.06]{orcid.pdf}\hspace{1mm}Gerhard Rigoll}\\
Technical University of Munich\\
Arcisstrasse 23, 80333 M\"unchen, Germany\\
	\texttt{\href{Martin.Knoche@tum.de}{Martin.Knoche@tum.de}} \\
}

\renewcommand{\undertitle}{A Preprint \\
Published 18th International Conference on Machine Vision Applications, Hamamatsu, Japan 2023, \copyright 2023 Institute of Electronics, Information and Communication Engineers (IEICE) \\
DOI: \url{https://doi.org/10.23919/MVA57639.2023.10216168}\vspace{-0.4cm}}
\renewcommand{\shorttitle}{Face Verification Human-Machine Fusion}

\hypersetup{
pdftitle={human-machine-fusion},
pdfsubject={},
pdfauthor={Martin~Knoche, Gerhard~Rigoll},
pdfkeywords={face, recognition, verification, artificial, intelligence, facial, confidence, score, platform, survey, human, matching, fusion, survey},
}

\makeatletter
\DeclareRobustCommand\onedot{\futurelet\@let@token\@onedot}
\def\@onedot{\ifx\@let@token.\else.\null\fi\xspace}
 
\def\ie{\emph{i.e}\onedot} 
\def\cf{\emph{cf}\onedot}

\def\etal{\emph{et al}\onedot}

\makeatother

\makeatletter
\providecommand*{\input@path}{}
\g@addto@macro\input@path{{./figures/}{./contents/}{./tables/}}
\makeatother

\usepackage[capitalize, nameinlink]{cleveref}
\crefname{section}{Sec.}{Secs.}
\Crefname{section}{Sec.}{Secs.}
\crefname{subsection}{Subsec.}{Subsecs.}
\Crefname{subsection}{Subsec.}{Subsecs.}
\Crefname{table}{Table}{Tables}
\crefname{table}{Table}{Tables}
\Crefname{figure}{Fig.}{Figs.}
\crefname{figure}{Fig.}{Figs.}
\Crefname{equation}{Equation}{Equations}
\crefname{equation}{Equation}{Equations}

\begin{document}
\maketitle

\setlength{\abovedisplayskip}{9pt}
\setlength{\belowdisplayskip}{9pt}

\begin{abstract}
	Nowadays, face recognition systems surpass human performance on several datasets. However, there are still edge cases that the machine can't correctly classify. This paper investigates the effect of a combination of machine and human operators in the face verification task. First, we look closer at the edge cases for several state-of-the-art models to discover common datasets' challenging settings. Then, we conduct a study with 60 participants on these selected tasks with humans and provide an extensive analysis. Finally, we demonstrate that combining machine and human decisions can further improve the performance of state-of-the-art face verification systems on various benchmark datasets. Code and data are publicly available on GitHub\footnote{\url{https://github.com/Martlgap/human-machine-fusion}}.
 \end{abstract}
 
 \section{Introduction}
In recent years, face recognition technology has become increasingly prevalent in various fields, such as security, surveillance, and healthcare. The continuous performance improvement can be primarily attributed to the development of deep learning models. These models have reached near saturation levels on standard benchmark datasets like LFW \cite{huang2008labeled}. However, several challenges remain to be addressed. For instance, the performance of face recognition systems is highly dependent on the quality of input images, as demonstrated by \cite{knoche2022susceptibility}, and is susceptible to pose~\cite{deng2018cplfw} or age~\cite{deng2017calfw} variations. More recently, the COVID-19 pandemic highlighted that face masks drastically affect the accuracy of traditional face recognition systems~\cite{wang2022mlfw}. Despite ongoing advancements, current face recognition systems are still need to be robust enough to achieve 100\% accuracy and thus require supervision by human operators in real-world applications. In this context, the human expert is a so-called ``super-recognizer'', making decisions based on their knowledge and the system's output.

Inspired by the recent work of Zheng \etal~\cite{zheng2022deep}, which proposes a fusion of a neural network with pathologists for predicting EBVaGC from histopathology, we aim to propose and analyze a similar fusion of state-of-the-art deep learning networks with human decisions for face verification. In contrast to Carragher's\etal~\cite{carragher2022simulated} previous work, our focus is on prioritizing machine decisions over human decisions.

In this work, we explore whether there are simple problems humans can solve but remain challenging for machines. To this end, we examine common face recognition benchmark datasets, focusing on cases where state-of-the-art models struggle. We hypothesize that humans can accurately and confidently solve the selected face verification problems that are difficult for machines. To test this hypothesis, we first conduct a baseline study with 60 participants to assess the consistency of human performance in face verification tasks without prior experience. Next, we carried out an investigation involving the same participants, asking them to verify selected image pairs that state-of-the-art face recognition models find challenging and to rate their confidence in their decisions.

Our main contributions are as follows:
\begin{itemize}
\item We analyze the performance of state-of-the-art face verification models on several standard benchmark datasets by examining specific image pairs where the models fail or exhibit low confidence in their decisions.
\item We conducted an in-depth analysis of a survey involving 60 participants, focusing on selected image pairs and highlighting the differences between machine and human performance, taking into account the participants' confidence levels.
\item We demonstrate that a clever combination of machine and human decisions can enhance the overall performance of face verification systems on various benchmark datasets.
\end{itemize}

 \section{Related Work}
Human performance in face verification or face matching has been extensively studied. However, only a few works have investigated human accuracy on popular datasets commonly used to evaluate automatic face recognition systems. In~\cite{kumar2009attribute}, the authors conducted a study using Amazon Mechanical Turk workers and evaluated human accuracy on the LFW~\cite{huang2008labeled} database, achieving 99.2\%. To the best of our knowledge, no studies have yet been conducted on human accuracy for other large benchmark datasets.

The following studies delve more into the psychological aspects of face verification: 

Howard \etal~\cite{howard2020human} investigated the influence of prior face identity decisions on subsequent human judgments about face similarity. Their study revealed that prior identity decision labels alter volunteers' internal criteria for judging face pairs.

Fysh and Bindemann~\cite{fysh2018human} examined the impact of onscreen trial labels with consistent, inconsistent, or unresolved information on face-matching accuracy. Their experiments demonstrated that human face-matching decisions are influenced by onscreen identifications, leading to increased accuracy when information is correct and increased errors when information is misleading.

In~\cite{sandford2021unfamiliar}, the authors explored the potential improvement in accuracy resulting from using multiple-image arrays in unfamiliar face-matching tasks. The study found that simultaneous viewing of multiple images of the same person improved matching accuracy.

Other works~\cite{white2022gfmt2, stantic2022oxford, abudarham2021face} have focused on developing standardized tests for human analysis in face-matching tasks.

 \section{Method}\label{sec:method}

\subsection{Survey Construction}\label{sec:dataset}
For constructing the surveys in our study, we utilize various modalities encompassed within commonly used face verification benchmark datasets, such as cross-age (CALFW)\cite{deng2017calfw}, cross-pose (CPLFW)\cite{deng2018cplfw}, cross-quality (XQLFW)\cite{knoche2021cross}, and masks (MLFW)\cite{wang2022mlfw}. All datasets contain $3\,000$ genuine (same identity) and $3\,000$ imposter (different identity) color image pairs, with a physical image resolution of $112\,\times\,112$ pixels.

First, we obtain predictions from three state-of-the-art face recognition models: 1) ArcFace*, a network based on ResNet50~\cite{he2016deep} and trained with additive angular margin loss~\cite{deng2022arcface}. 2) FaceTrans*, a vision transformer architecture trained on faces~\cite{zhong2021facetransformer}. We use Knoche's\etal~\cite{knoche2023octuplet} publicly available and more robust fine-tuned variants of both models. 3) ProdPoly, a polynomial neural network architecture trained on faces proposed by Chrysos \etal~\cite{chrysos2020p}. We also calculate their certainty scores using the method proposed by Knoche \etal~\cite{knoche2023explainable}.

Next, we divide the mean certainty of all models, ranging from 0 to 100\%, into five equally sized bins. We then randomly select one imposter and one genuine image pair from each bin and each dataset, resulting in $40$ image pairs. This collection serves as the pool of image pairs for our baseline survey, which we will denote as \survey{1}.

Second, to construct \survey{2--5}, we divide the mean certainty of all models into ten equally sized bins and randomly select 30 samples from each bin to create a pool. Additionally, we identify all image pairs from the abovementioned databases where the minimum certainty of all three models falls below a threshold of 50\% and include them in the pool. This procedure results in four pools containing $7\,445$ image pairs. \Cref{tab:dataset_stats} summarizes the image pair statistics for all constructed surveys.

\begin{table}[t]
    \centering
    \caption{Detailed image pair statistics for all surveys.}
\begin{tabular}{lccccc}
    \toprule
          & \multicolumn{5}{c}{Dataset} \\
    \cmidrule{2-6}      & \cl BASE & CALFW & \cl CPLFW & MLFW  & \cl XQLFW \\
    \midrule
    Survey & \cl 1 & 2     & \cl 3 & 4     & \cl 5 \\
    \midrule
    \# Pairs in Pool & \cl 40 & 684   & \cl 1385 & 3561  & \cl 1815 \\
    \# Pairs per User & \cl 40 & 12    & \cl 24 & 60    & \cl 31 \\
    \% of Total Pairs & \cl  -  & 5.37\% & \cl 10.28\% & 28.58\% & \cl 13.58\% \\
    \% Genuine & \cl 50.00\% & 47.08\% & \cl 44.55\% & 48.16\% & \cl 44.90\% \\
    \bottomrule
    \end{tabular}%
    \label{tab:dataset_stats}
\end{table}

\subsection{Study Design}\label{sec:design}
Each of our $60$ participants must register for our online study via a website, where they can provide their age, ethnicity, and gender. After registration, participants can complete the survey at their convenience. We have \survey{1--5} available for participants to complete.

The baseline \survey{1} is identical for all participants, \ie, all $40$ image pairs are displayed to every user in the same order.

All subsequent surveys are personalized for each participant. The image pairs for \survey{2--5} are randomly selected from the corresponding pool of images (see \cref{sec:dataset}) and are unique, meaning no image pair is displayed twice. The ratio of genuine and imposter pairs varies for each user. Every user is asked to answer $167$ image pairs, resulting in $9\,845$ questions. The last three participants received fewer image pairs as the respective image pair pools were depleted.

Participants are not informed of the ratio of genuine and imposter pairs or the labels of the image pairs. To prevent any negative influence on the participants' performance, we do not display the models' predictions to the participants as proposed in~\cite{fysh2018human}. We also do not display the outcome of previous questions to the participants. This could influence their decision on subsequent questions, particularly by revealing whether the previous answer was correct or incorrect~\cite{howard2020human}.

For each image pair in the survey datasets, participants need to decide whether the two images show the same or different identities and rate their certainty regarding the answer to the first question. Participants adjust a slider, which ranges from ``uncertain'' to ``certain'' without any numerical indicators or additional information. The default position of the slider is at the midpoint, corresponding to a value of $0.5$, with the internal range of values extending from 0 (uncertain) to 1 (certain). The duration is also recorded.

After completing at least one question in the survey, participants can leave the survey and review the accuracies of their answers thus far, along with the corresponding FaceTrans* model's performance.

The survey platform's code was written in Python using the Flask framework and hosted on Google Cloud. 



\begin{figure}[b]
    \centering
    \includegraphics[width=0.6\columnwidth]{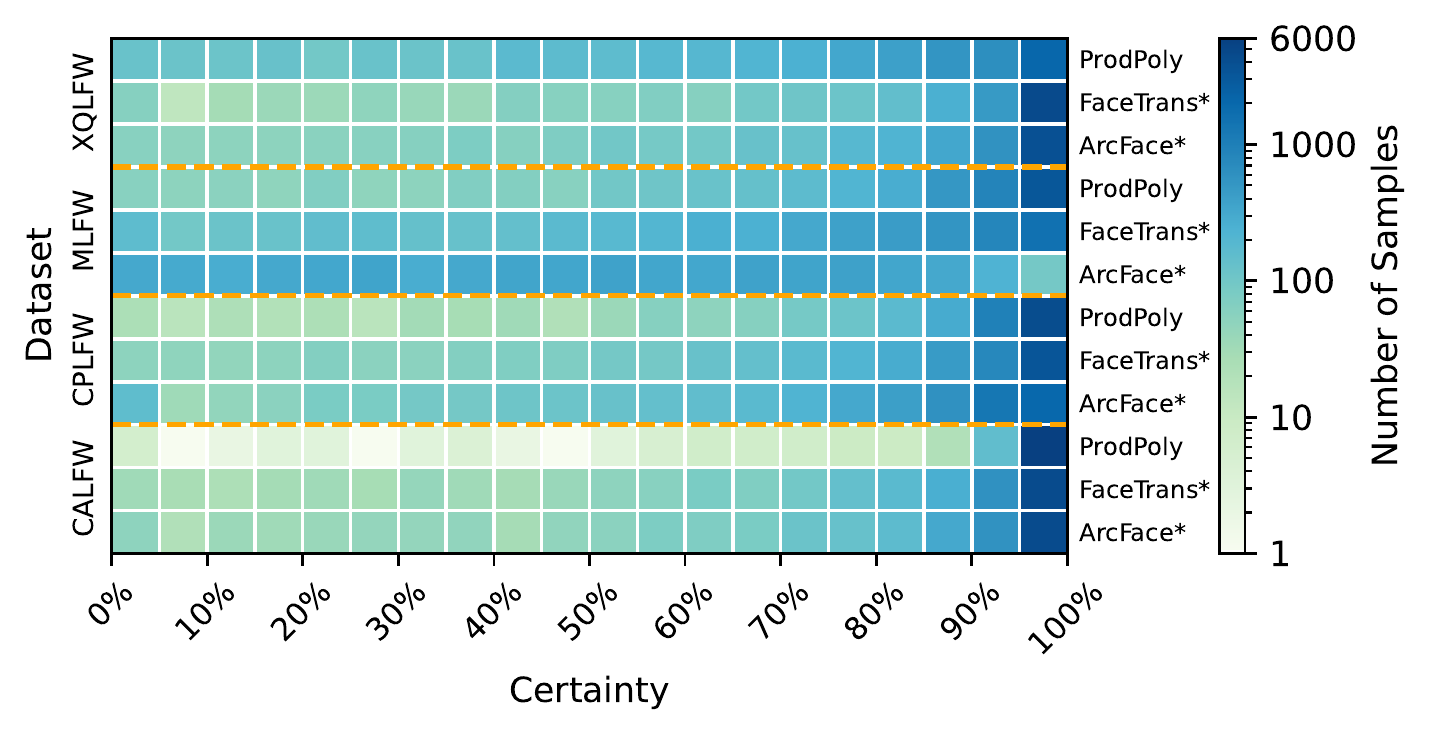}
    \caption{Certainty distributions on standard datasets for face recognition networks.}
    \label{fig:conf_scores}
\end{figure}

\subsection{Human-Machine Fusion}
\label{sec:fusion}
In this work, we examine human-machine decision fusion in face verification. Our fusion algorithm can be outlined: We set a minimum certainty threshold $t_{\text{min}}$ for machine decisions to be directly accepted. Any machine decisions falling below this threshold are considered for comparison with the calibrated human certainty. If the human certainty for a particular question exceeds a specified threshold $t_{\text{max}}$ and surpasses the machine certainty by at least $d_{\text{min}}$, we accept the human decision. Otherwise, we maintain the machine's decision. After collecting human feedback on the pool of image pairs for all datasets, the proposed fusion algorithm is applied to all $6\,000$ pairs in each dataset.



 \section{Results}\label{sec:results}

\subsection{Dataset Insights}\label{sec:insights}

In~\cref{fig:conf_scores}, we present the distribution of certainty scores for our three models on the four datasets (see~\cref{sec:dataset}). For all datasets, it is evident that the machine predicts with very high certainty ($\ge 95\%$) most of the time. The masked dataset (MLFW) appears to be the most challenging for the models, which is also reflected in the accuracy (see~\cref{tab:accuracies}). Additionally, the number of image pairs with certainty $\le 50\%$ is the highest in this dataset (refer to~\cref{tab:dataset_stats}).

\begin{figure}[t]
    \centering
    \includegraphics[width=0.7\columnwidth]{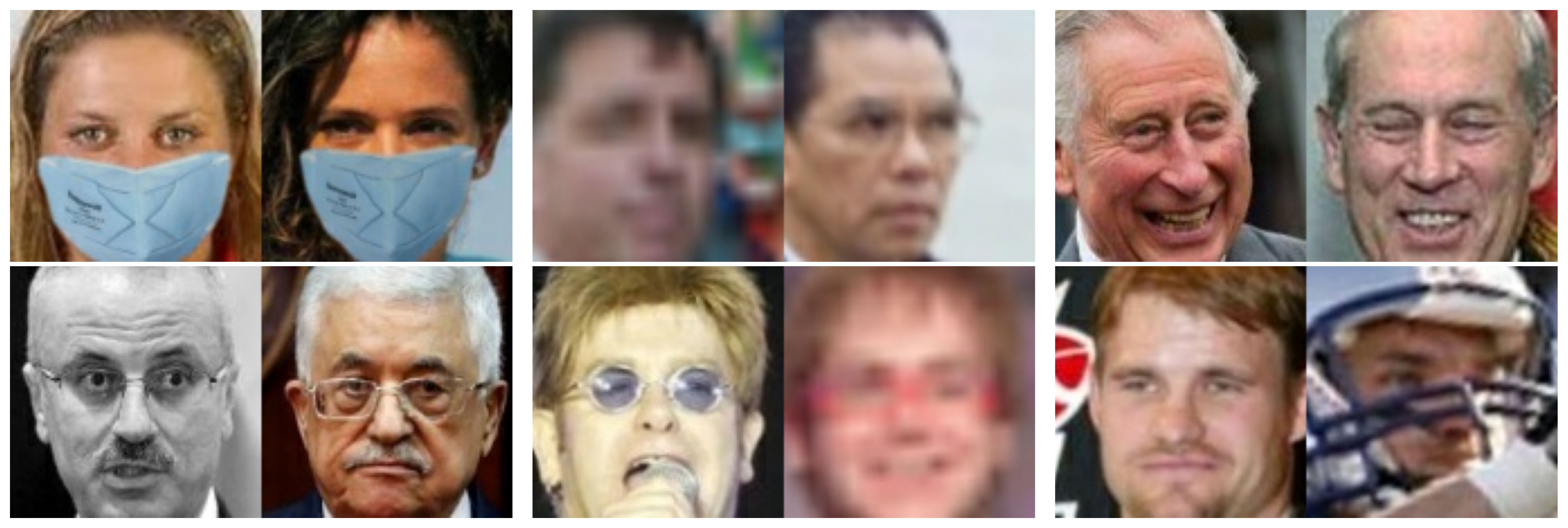}

    \caption{Example image pairs with correct human decisions and higher certainty than machine.}
    \label{fig:findpairs}

\end{figure}

Furthermore, we conduct a closer visual inspection of the image pairs in the datasets, selecting pairs where the machine prediction is incorrect and has low confidence. In contrast, the human prediction is correct with very high confidence. In~\cref{fig:findpairs}, we showcase three imposter and three genuine image pair examples that meet these criteria. Human operators made correct decisions for all these samples and rated them with 100\% certainty, while the machine was incorrect and had much lower confidence.

\subsection{Participants Characteristics}\label{sec:participants}
\vspace{-0.2cm}
\begin{figure}[b]
    \centering
    \includegraphics[width=0.7\columnwidth]{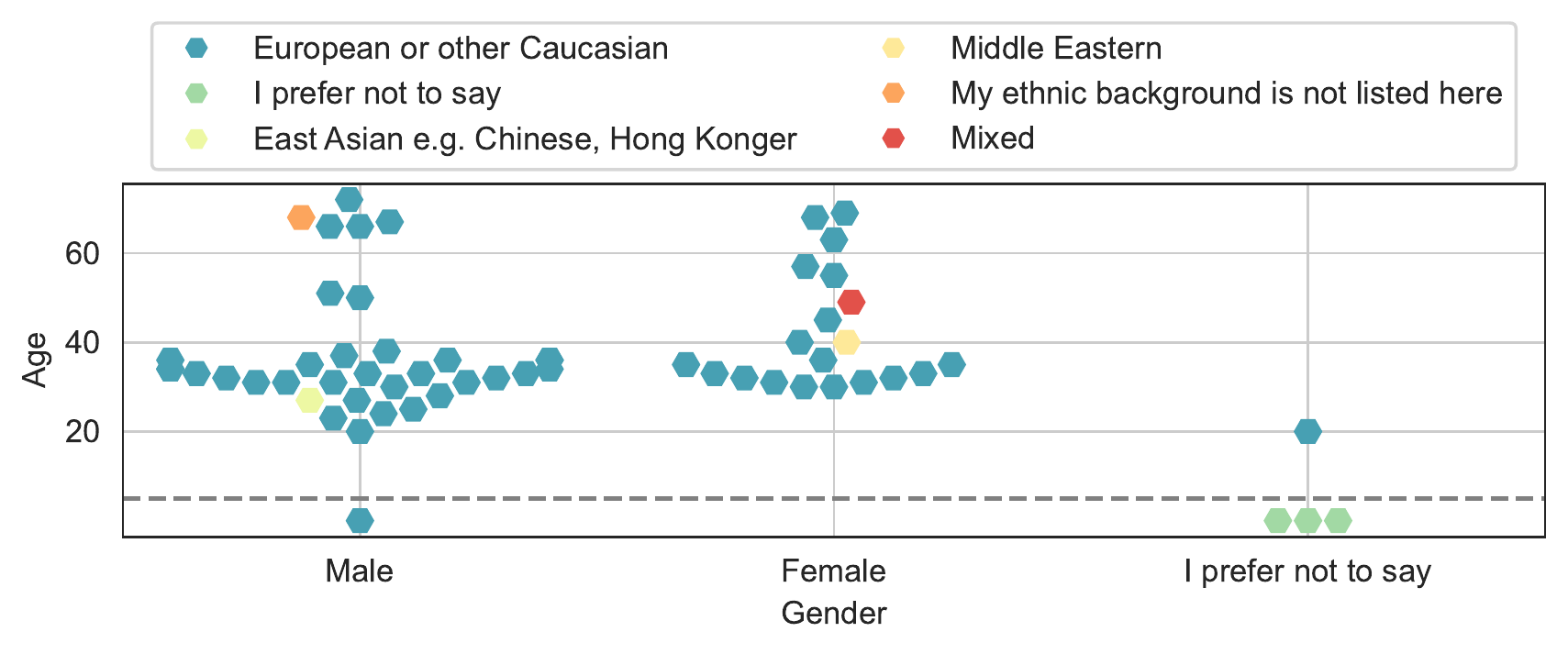}
    \caption{Age, gender and ethnicity distribution of the participants.}
    \label{fig:user_stats}
\end{figure}

The data for this study was collected online over 14 days in March/April 2023. \Cref{fig:user_stats} displays the age, gender, and ethnicity distribution of the participants. The mean age is $37.22$ (std=$17.41$, min=20, max=72), with 60\% of the volunteers being male, 33.33\% female, and four participants preferring not to state their gender. We primarily recruited European/Caucasian users, which, although it may limit the generalizability of the study, more closely reflects the ethnicity distribution of the datasets used in our research, as reported by~\cite{wang2019racial}.

\subsection{Experiment 1}\label{sec:exp1}

\begin{figure}[t]
    \centering
    \subfigure{
        \includegraphics[width=1\columnwidth]{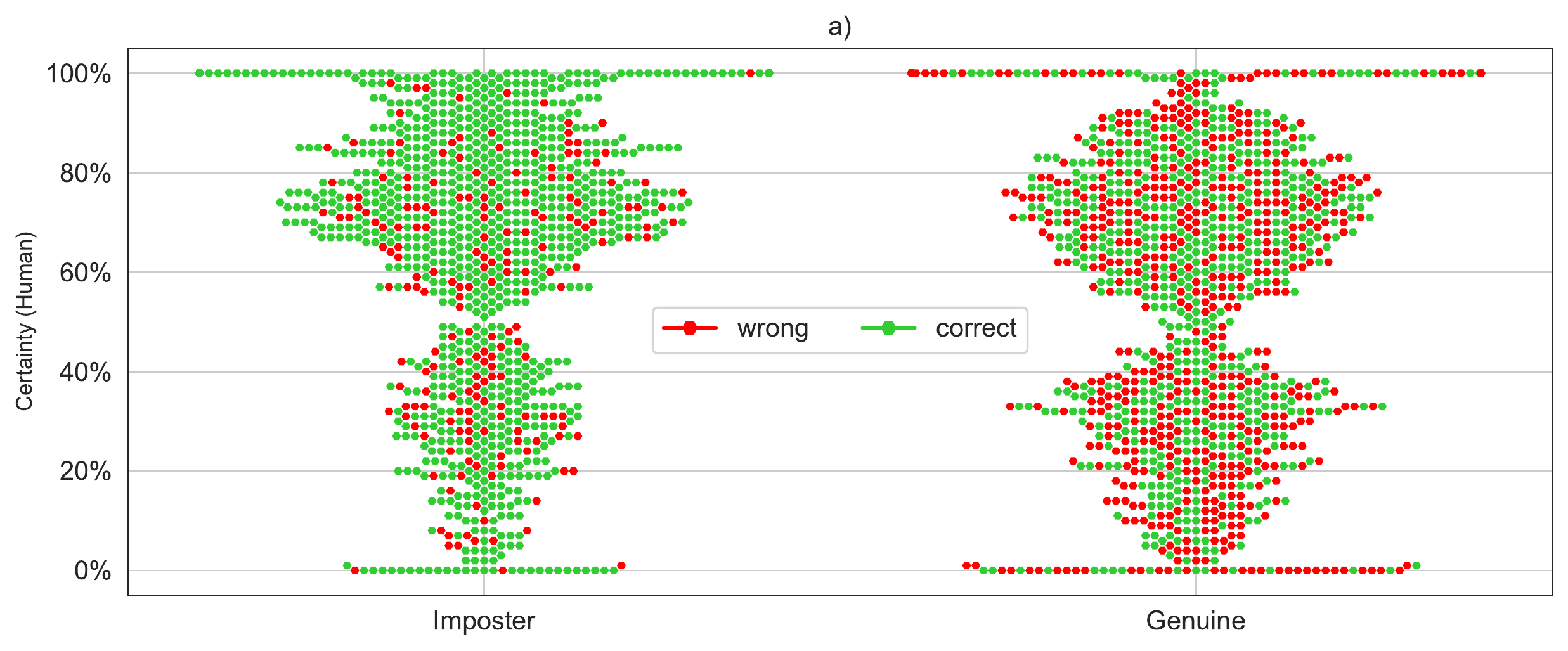}\label{fig:base_swarmplot}
    }
    \subfigure{
        \includegraphics[width=0.48\columnwidth]{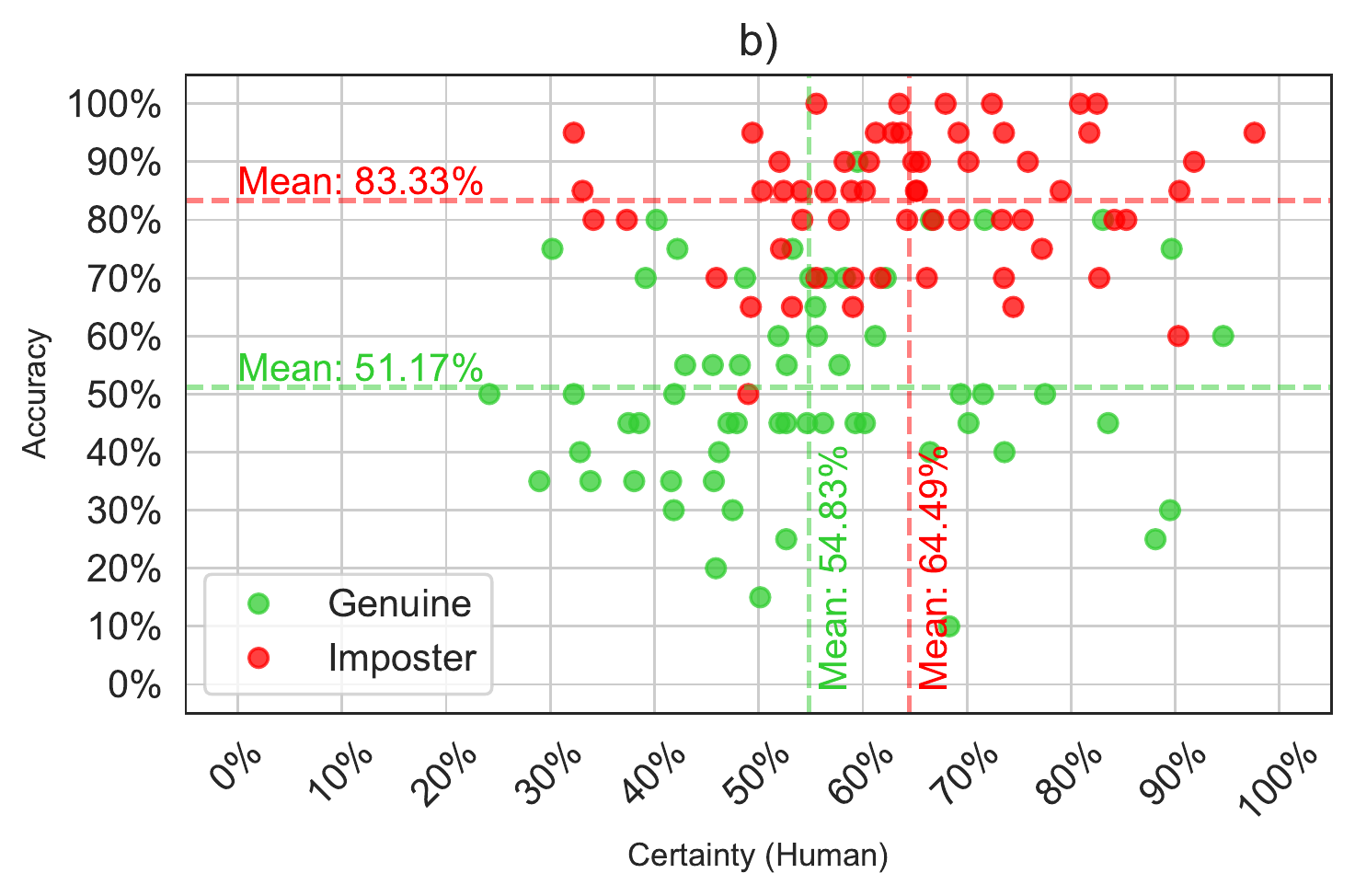}\label{fig:base_acc_cert_p_user}%
    }
    \subfigure{
        \includegraphics[width=0.48\columnwidth]{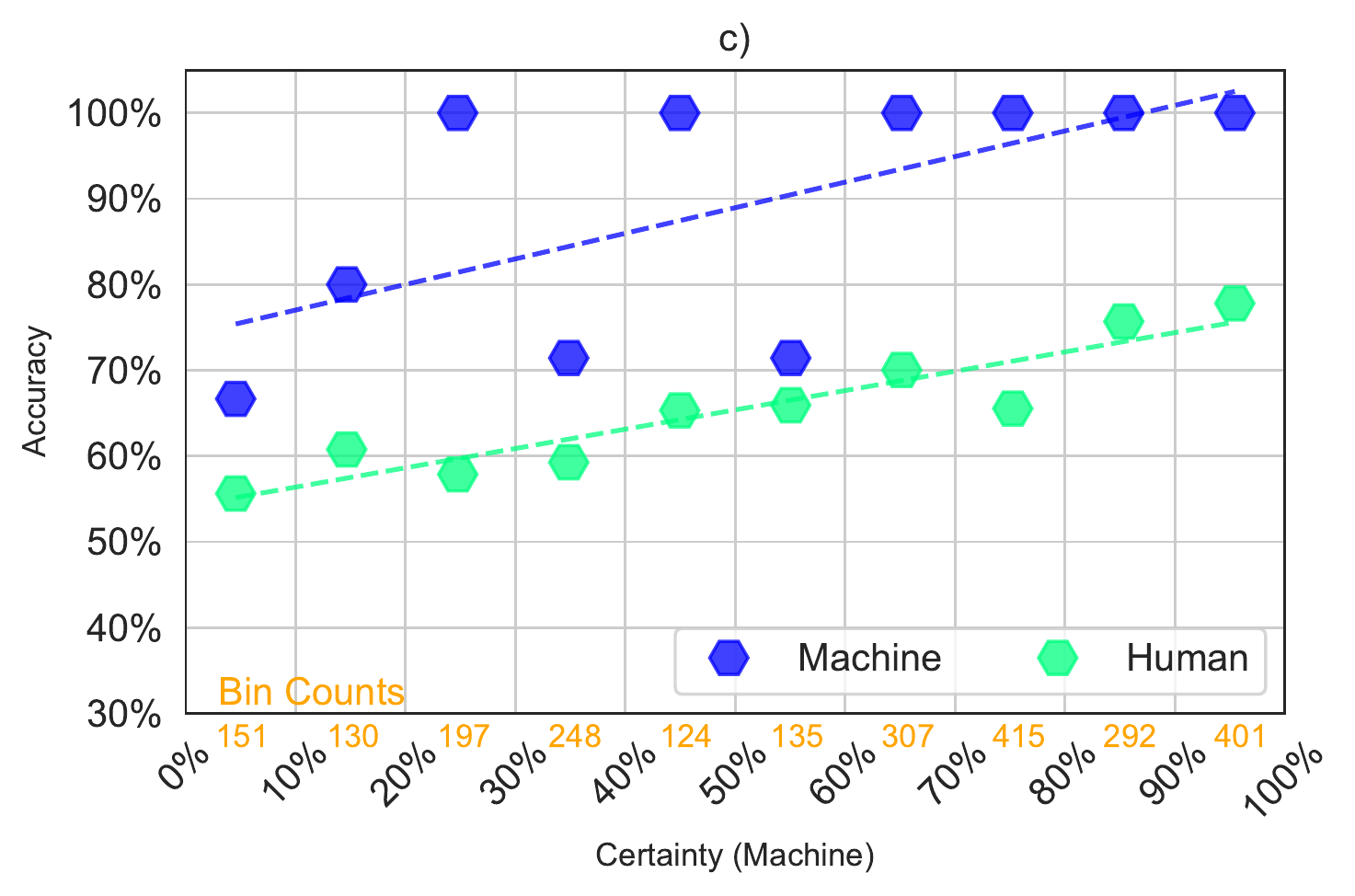}\label{fig:base_acc_cert_hist} 
    }
    \caption{a) User prediction distribution b) Mean certainty and accuracy per user. c) Accuracy comparison for machine certainty histogram bins.}
    \label{fig:both}
\end{figure}

The collected feedback from \survey{1} is summarized in~\cref{fig:base_swarmplot}, which reveals that humans perform better on imposter pairs than on genuine pairs. This observation is also evident in~\cref{fig:base_acc_cert_p_user}, displayed using an alternative representation. Here, it can be seen that both accuracy and certainty are higher for imposter pairs than for genuine pairs. Additionally, a correlation factor of 0.44 demonstrates a moderate correlation between the accuracy and certainty of the participants' answers. A notable dip at the 50\% certainty level is visible in~\cref{fig:base_swarmplot}, which can be attributed to the rating slider design. The slider is initially centered for each question and must be moved by the user. We observe slightly more answers with certainties above 50\%, particularly for imposter pairs, resulting in a total mean certainty of 59.66\% (std=14.61) and a mean accuracy of 67.25\% (std=8.53) for \survey{1}. 

We also assessed the average user certainty for true positives (56.73\%, std=27.74), false positives (55.63\%, std=25.73), true negatives (66.27\%, std=28.94), and false negatives (52.83\%, std=30.33) image pairs. The highest certainty is achieved for true negatives, consistent with the results in~\cref{fig:base_acc_cert_p_user}. \Cref{fig:base_acc_cert_hist} illustrates the mean accuracies for each machine-certainty bin. We compare human performance with the mean of three machine models (\cf~\cref{sec:dataset}). As expected, the machine ensemble outperforms human operators, primarily because we selected image pairs covering the entire range of machine certainties, acknowledging that machines excel over humans in most cases. The number of image pairs assessed within each bin increases with certainty and accuracy.

\begin{figure}[t]
    \centering
    \includegraphics[width=1\columnwidth]{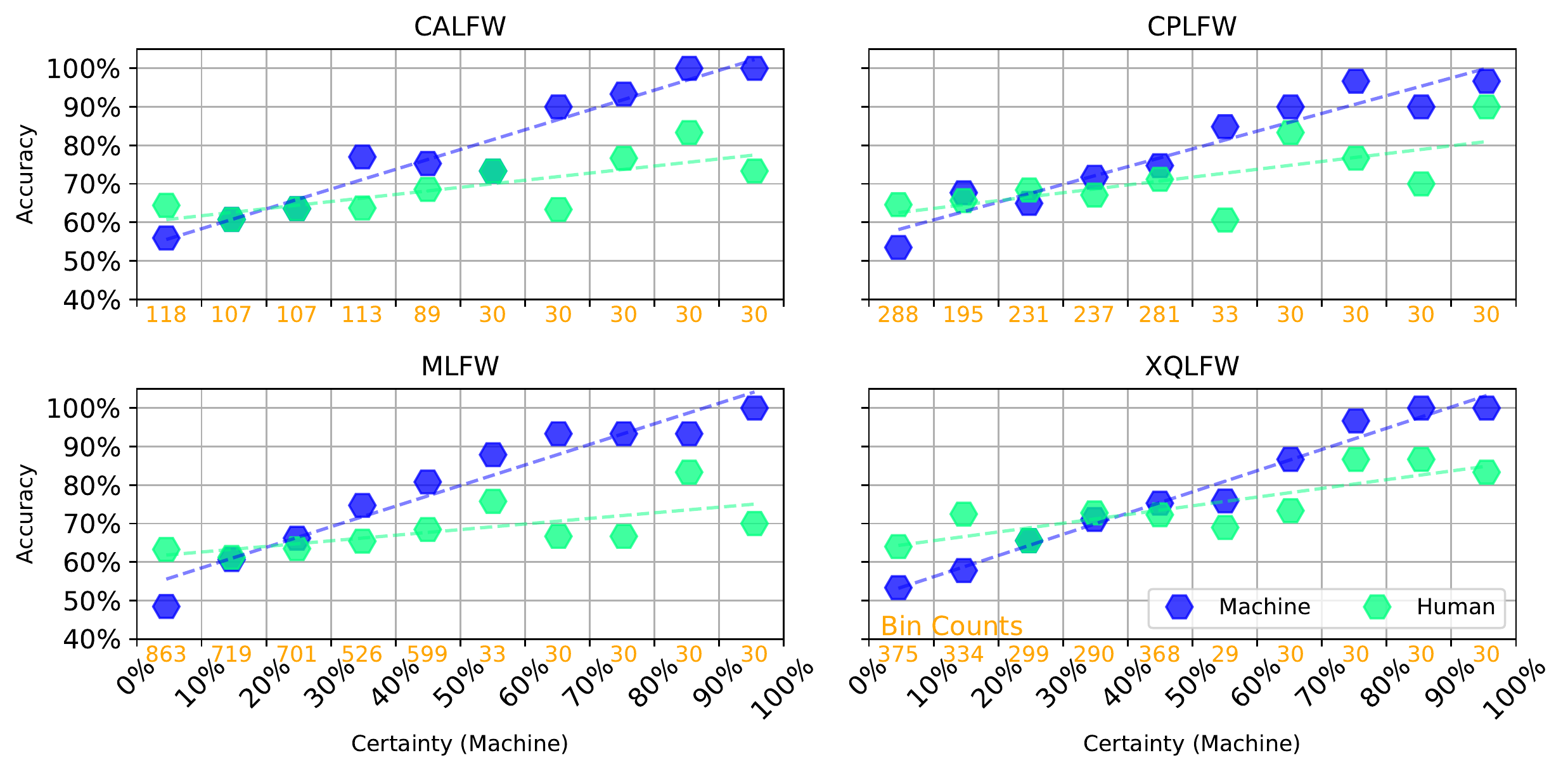}
    \caption{Accuracy comparison for machine certainty histogram bins across the datasets.}
    \label{fig:acc_cert_histogram}
\end{figure}

Interestingly, the decision duration (mean=10.71 sec, std=8.38) varied significantly (p=0.00003) based on the user's prediction. Users took an average of 10.22 (std=8.12) sec for the ``different'' answer and 11.66 (std=8.78) sec for the ``same'' answer. Investigating the precise reasons behind this finding would be an intriguing endeavor.

Lastly, accuracy analysis concerning participants' ages revealed no significant differences.

\subsection{Experiment 2}\label{sec:exp2}

\Cref{fig:acc_cert_histogram} demonstrates that the mean accuracies for humans are higher than those for the machine ensemble when the machine ensemble certainties are very low (less than 10\%) in \survey{2--5}. This supports our fundamental assumption and motivation for our fusion approach (see~\cref{sec:fusion}). The accuracy for both human and machine ensembles increases with rising certainty. The number of human decisions increases as the certainty score decreases. 

\begin{table}[b]
    \centering
    \caption{Face verification accuracy [\%] comparison and number of human decisions in fusion-algorithm.}
\begin{tabular}{rcccc}
    \toprule
          & \multicolumn{4}{c}{Dataset} \\
    \cmidrule{2-5}\multicolumn{1}{l}{Model} & \cl CALFW~\cite{deng2017calfw} & CPLFW~\cite{deng2018cplfw} & MLFW~\cite{wang2022mlfw} & XQLFW~\cite{knoche2021cross} \\
    \midrule
    \multicolumn{1}{l}{ArcFace*~\cite{knoche2023octuplet}} & \cl 93.85 & 88.37 & 73.53 & 93.27 \\
     + Human & \cl 94.03 \smaller{(\clg{+0.18})} & 88.83 \smaller{(\clg{+0.47})} & 75.88 \smaller{(\clg{+2.35})} & 93.65 \smaller{(\clg{+0.38})} \\
    \# Human Decisions & \cl 241 \smaller{(4.02 \%)} & 470 \smaller{(7.83 \%)} & 1871 \smaller{(31.18 \%)} & 257 \smaller{(4.28 \%)} \\
    \midrule
    \multicolumn{1}{l}{FaceTrans*~\cite{knoche2023octuplet}} & \cl 94.93 & 91.58 & 85.63 & 95.12 \\
     + Human & \cl 94.93 \smaller{(\clg{+0.00})} & 91.73 \smaller{(\clg{+0.15})} & 85.70 \smaller{(\clg{+0.07})} & 95.28 \smaller{(\clg{+0.17})} \\
    \# Human Decisions & \cl 187 \smaller{(3.12 \%)} & 311 \smaller{(5.18 \%)} & 778 \smaller{(12.97 \%)} & 172 \smaller{(2.87 \%)} \\
    \midrule
    \multicolumn{1}{l}{ProdPoly~\cite{chrysos2020p}} & \cl 96.03 & 92.75 & 91.30 & 86.90 \\
     + Human & \cl 96.13 \smaller{(\clg{+0.10})} & 92.82 \smaller{(\clg{+0.07})} & 91.35 \smaller{(\clg{+0.05})} & 88.05 \smaller{(\clg{+1.15})} \\
    \# Human Decisions & \cl 17 \smaller{(0.28 \%)} & 125 \smaller{(2.08 \%)} & 325 \smaller{(5.42 \%)} & 605 \smaller{(10.08 \%)} \\
    \bottomrule
    \end{tabular}%
    \label{tab:accuracies}%
\end{table}

\subsection{Human-Machine Fusion}\label{sec:fusion}

In~\cref{tab:accuracies}, we provide an overview of the performance of the three state-of-the-art models utilized in this study, combined with our fusion technique, on the four face verification benchmark datasets. The parameters $t_{\text{min}}$, $t_{\text{max}}$, and $d_{\text{min}}$ were empirically set to $0.37$, $0.11$, and $0.05$, respectively. The results demonstrate that our fusion approach surpasses the individual models on all datasets. Furthermore, it is apparent that the lower the model's accuracy, the more human decisions are required to enhance the overall performance of the fusion approach.

\subsection{Limitations}\label{sec:limitations}

As mentioned in~\cref{sec:participants}, we recruited only 60 participants, primarily of European/Caucasian ethnic background, which may limit the generalizability of our study~\cite{phillips2014comparison}. Our study encompasses a combination of familiar and unfamiliar face matching, which could also influence generalizability.~\cite{johnston2009familiar}.

 \section{Conclusion}\label{sec:conclusion}
\vspace{-0.2cm}
Having initially analyzed the distribution of certainty for machine models across various datasets, we constructed five surveys from these datasets and evaluated them with 60 participants. The key findings of the results are that humans perform better on imposter image pairs than on genuine pairs. Additionally, users have very high certainty and provide the correct answer for some pairs where the machine has low certainty. Ultimately, our fusion algorithm demonstrates that combining machine and human models can further improve machine accuracies.

In the future, we plan to extend our work by calibrating user certainties to further enhance human-machine fusion accuracy. We also intend to investigate the effect of presenting face verification explanation maps from machine models to users.

{\small
\bibliographystyle{IEEEbib}
\bibliography{main}
}

\end{document}